\documentclass[conference,fleqn]{IEEEtran}
\IEEEoverridecommandlockouts
\usepackage{amsmath,amssymb,amsfonts}
\usepackage{algorithmic}
\usepackage{graphicx}
\usepackage{textcomp}
\usepackage{graphics} % for pdf, bitmapped graphics files
\usepackage{epsfig} % for postscript graphics files
\usepackage{booktabs}
\usepackage[compress]{cite}
\usepackage[font={footnotesize}]{caption}
\usepackage[font={footnotesize,up,singlespacing}]{subcaption}
\usepackage[switch,pagewise]{lineno}
\usepackage[table,xcdraw]{xcolor}
\usepackage[utf8]{inputenc}

\graphicspath{ {./images/} }

\title{\LARGE \bf
Dynamic Vision Sensor integration on FPGA-based CNN accelerators for high-speed visual classification
}

\author{ \parbox{3 in}{\centering Alejandro Linares-Barranco*, Antonio Rios-Navarro, Ricardo Tapiador-Morales**
         \thanks{*This work is supported by Samsung funded project NPP-phase2, coordinated by Institute of Neuroinformatics, and by the excellence project from the Spanish government grant (with support from the European Regional Development Fund) COFNET (TEC2016-77785-P)}
         \thanks{**R.  Tapiador  has  been  supported  by  a "Formación de Personal Investigador" Scholarship from the University of Seville.}\\
         Robotic and Tech of Computers Lab. University of Seville. SPAIN\\
         {\tt\small alinares@us.es}}
         \hspace*{ 0.5 in}
         \parbox{3 in}{ \centering Tobi Delbruck\\
         Institute of NeuroInformatics. University of Zurich. Switzerland\\
         {\tt\small tobi@ini.uzh.ch}}
}

\begin{document}

\maketitle
%\thispagestyle{empty}
%\pagestyle{empty}

%%%%%%%%%%%%%%%%%%%%%%%%%%%%%%%%%%%%%%%%%%%%%%%%%%%%%%%%%%%%%%%%%%%%%%%%%%%%%%%%
\begin{abstract}

Deep-learning is a cutting edge theory that is being applied to many fields. For vision applications the Convolutional Neural Networks (CNN) are demanding significant accuracy for classification tasks. Numerous hardware accelerators have populated during the last years to improve CPU or GPU based solutions. This technology is commonly prototyped and tested over FPGAs before being considered for ASIC fabrication for mass production. The use of commercial typical cameras (30fps) limits the capabilities of these systems for high speed applications. The use of dynamic vision sensors (DVS) that emulate the behaviour of a biological retina is taking an incremental importance to improve this applications due to its nature, where the information is represented by a continuous stream of spikes and the frames to be processed by the CNN are constructed collecting a fixed number of these spikes (called events). The faster an object is, the more events are produced by DVS, so the higher is the equivalent frame rate. Therefore, these DVS utilization allows to compute a frame at the maximum speed a CNN accelerator can offer. In this paper we present a VHDL/HLS description of a pipelined design for FPGA able to collect events from an Address-Event-Representation (AER) DVS retina to obtain a normalized histogram to be used by a particular CNN accelerator, called NullHop. VHDL is used to describe the circuit, and HLS for computation blocks, which are used to perform the normalization of a frame needed for the CNN. Results outperform previous implementations of frames collection and normalization using ARM processors running at 800MHz on a Zynq7100 in both latency and power consumption. A measured 67\% speed-up factor is presented for a Roshambo CNN real-time experiment running at ~160fps peak rate.

\end{abstract}

%%%%%%%%%%%%%%%%%%%%%%%%%%%%%%%%%%%%%%%%%%%%%%%%%%%%%%%%%%%%%%%%%%%%%%%%%%%%%%%%
\section{INTRODUCTION}

Field-Programmable-Gate-Arrays (FPGA) have demonstrated superior impact not only in prototyping for deep-learning architectures, but also for commercial products in this field. For visual applications, the convolutional neural networks (CNNs) represent one of the most utilised approaches for a number of continuously increasing large-scale machine vision tasks~\cite{Krizhevsky_etal12,He_etal15a,LeCun_etal15}. A CNN has an architecture composed of several layers of features extraction. First layer is composed of a number of convolutional filters that extracts several basic features from the input image. Consecutive deeper layers are also composed of a number of convolutional filters, but in this case they take as input a set of extracted features from the previous layer (that could be all of them) for more complex feature extractions. The weights used for these convolution operations are obtained through a training method. The core operation of a CNN layer is the convolution expressed in (\ref{eq:conv}), where $F^{out}_{(a,b)}$ is an output pixel of a convolution applied to a feature in $F^{in}$ with a kernel $C$. To obtain the final feature output, the selected input features, computed with their corresponding kernels, are accumulated in the output feature. It is normal to find non-linear operation to connect two layers, like ReLU (Rectification Unit), and pooling operations to reduce the domain problem in a closer set of complex features that can be classified by a fully-connected-layer at the end of the architecture.

\begin{equation}
F^{out}_{(a,b)} = \sum\limits_{\tiny i=-N_{in}/2}^{N_{in}/2} C_{i,j}* F^{in}_{(a-i/2, b-i/2)}
\label{eq:conv}
\end{equation}

Their computational simple core concept merged with efficient supervised training techniques have made them the main selected method for image features extraction at high-level semantic for classification, localization, and detection visual processing tasks ~\cite{Sermanet_etal13}.

CNN are typically trained using back-propagation technique, with a relatively well sized set of labeled examples, to allow it to inference the correct class for an input image not used during the training. The network is usually trained on hardware platforms such as
graphical processing units (GPUs) or specialized architectures, as in~\cite{Jouppi2017} previously to the use of the CNN for a particular application. The training process is computationally heavy and it requires several iterations and fine tuned datasets to extract the best accuracy from a fixed CNN architecture. 

After training, the CNN is used for runtime inference. Depending on the CNN architecture, this inference could become computationally expensive. State of the art (SOA) CNNs typically require several billion multiply-accumulate (\textbf{MAC}) operations per image. Therefore, the use of mobile processors or mobile GPUs to run a CNN can become expensive in a power-constrained mobile platform and / or not feasible for particular high-speed applications. 

For instance, in \cite{ai-benchmark-arxiv-2018} the model Inception V3 for object recognition tasks has an accuracy of around 78\% with ~27M parameters and ~12GOP, requiring 50ms to compute a frame on a Kirin 980 smartphone processor\footnote{Taken from http://ai-benchmark.com/ranking on 25th March 2019}. The Inception ResNet V1 for face recognition has an accuracy of around 98\% and needs around 113ms in the same mobile chip. And the ICNET architecture for semantic segmentation, used for autonomous driving, has an accuracy of around 70\% being able to obtain real-time computation (30fps) with powerful GPUs installed on a commercial car, but it requires 275ms on the Kirin 980 chip. This last algorithm needs 64ms when being executed on a Free-TPU for FPGA. These examples highlight that FPGA solutions for solving real problems are getting closer and closer. 

However, the limit of 30fps of commercial cameras will be an expensive problem to solve if high-speed cameras start to be installed in vehicles. The Neuromorphic Engineering community has being working for more that 20 years on bio-inspired sensors (called Dynamic Vision Sensors - DVS) \cite{Lichtsteiner_etal08,Brandli_etal14} that offer equivalents $\approx 20k fps$ compared to CCD cameras. The principle behind is the fact of allowing each pixel to work independently respect to its neighbors for sending its current status if a feature has been detected and need to be highlighted. Most of these sensors are tracking local luminosity changes, at pixel level, in such a way that they behave as a neuron, sending a spike or event when the detected change has surpassed a threshold. Typically these events are sent with a polarity that tells whether the change has been from darker to lighter or reverse. In Fig \ref{fig:roshambo-scenario} (monitor box of laptop screen) it is shown a 2D histogram of 2k captured events for 6ms in this particular case. It can be seen how positive events (white) represent the front part of the hand in motion, while negative events (black) represent the back part of the same hand. The faster the hand, the smaller is the required time to obtain 2k events. If these histograms could be used for CNNs it means that fps can be adjusted to the particular speed of the scenario where the object has to be detected and / or classified.

In this paper we compare the performance for histogram collection and normalization in a pipelined High-Level-Synthesis(HLS) hardware implementation for FPGA, respect to a previously used embedded software approach for a CNN accelerator scenario.

The paper is structured as follows: 
Section II introduces the NullHop accelerator and its FPGA deployment considerations. Section III describe in details the circuit designed to collect and normalize a histogram of events to be inference by NullHop. Section IV show performance results and comparison respect to sofware implementations over a RoShamBo CNN example for the NullHop.

\section{NullHop Accelerator}
The CNN accelerator scenario selected takes advantage of the sparsity of networks (feature maps and kernels) for saving computation and memory accesses during inference. It is called \textit{NullHop} \cite{Aimar_etal2019}, which exploits activation sparsity by two main features: (1) its ability to skip over 
zeros (zero-skipping) in the input CNN layers without any wasted clock cycles and redundant MACs. (2) The compression scheme that is optimized for sparsely activated CNN layers. This compression reduces external memory accesses and is more efficient 
than other run-length encoding schemes~\cite{Chen_etal16}.
Similar to the current SOA accelerators~\cite{pham_neuflow:_2012,Du2015,Chen_etal16,Sim_etal16,Moons2016,Moons2017,Lee18,Yin2017,Shin2017}, NullHop uses a configurable processing pipeline that maintains high efficiency across a range of CNN kernel sizes and numbers of feature maps.

\subsection{Architecture}

\begin{figure}[ht!]
  \centering
  \includegraphics[width=1.0\linewidth]{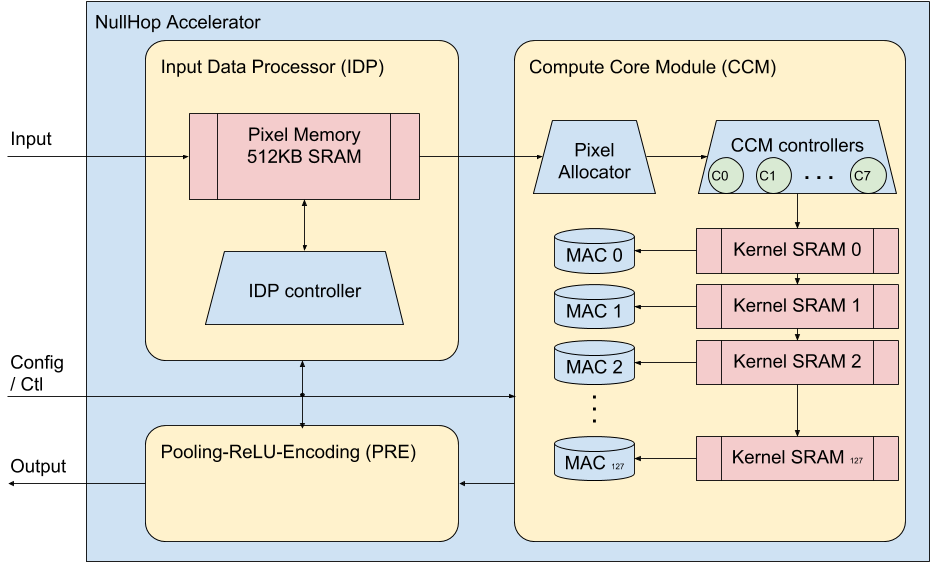} 
  \caption{NullHop CNN accelerator block diagram.}
\label{fig:nullhop}
\end{figure}

Figure \ref{fig:nullhop} shows the block diagram of the \textit{NullHop} accelerator. The interface allows to feed data through a 32-bit input data bus; read data from an output 32-bit bus; configure the accelerator parameters with an input configuration bus; and several control lines for clock, reset and bus handshake signals. The accelerator implements one convolutional stage with 128 MAC units followed by a ReLU transformation and then a max-pooling stage that can be optionally used. The CNN is implemented in a forward pass. Therefore, the accelerator evaluates convolutional stages sequentially. The input feature maps and the kernel values for the current convolutional layer are stored in two independent SRAM blocks through input and configuration interfaces. 

The output feature maps produced by the current layer are streamed off-chip through the output interface to be stored in the external memory. Then, they are  streamed back to the accelerator SRAM when the accelerator has finished processing the current layer. The feature maps are always stored in a compressed format that is never decompressed but rather decoded during the computation.

NullHop uses a sparse matrix compression technique, which produces an average compression level higher than other methods, such as in~\cite{Chen_etal16}. And it is easier to decode than the Huffman coding used in~\cite{Moons2016}. This coding uses a Sparsity Map (\textbf{SM}), that is a 3D mask with the same number of entries as number of pixels in the feature maps; and a Non-Zero Value List (\textbf{NZVL}). The SM is used to reconstruct the positions of the non-zero pixels that are present in the NZVL. 

The Input Decoding Processor (\textbf{IDP}) reads a small portion of the compressed input feature maps to pass non-zero pixels to the Compute Core Module (\textbf{CCM}). Therefore, the IDP is skipping zero pixels in the compressed input feature maps without wasting any MAC operation. In addition to the pixel values, the IDP also forwards the pixels' positions (row, column, and input feature map index) to the CCM. The Pixel Allocator (at CCM) allocates incoming pixels through available \textit{Controllers}. Each \textit{Controller} schedules operations for a subset of the MAC blocks and it submits the needed read requests to the corresponding Kernel SRAM. All MAC blocks under the same Controller receive the same input pixel from their Controller, but weights from different kernels, producing pixels in different output feature maps. The convolution results are sent through an optional ReLU transformation and a max-pooling stage before going to the pixel stream compression block (PRE module of the figure). The compressed output feature maps are then sent off-chip.
This NullHop accelerator has been described using SystemVerilog and it has been synthesized for both ASIC and FPGA, as described with details in \cite{Aimar_etal2019}.

\subsection{FPGA implementation}
\begin{figure}[ht!]
  \centering
  \includegraphics[width=1.0\linewidth]{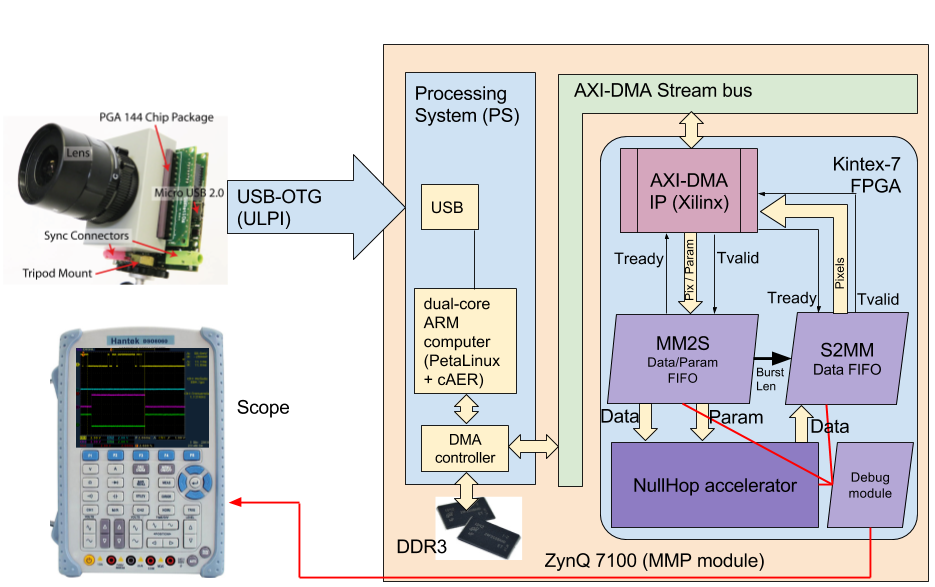} 
  \caption{NullHop integration on PSoC block diagram.}
\label{fig:nullhop-axi}
\end{figure}

The NullHop accelerator has been deployed into a PSoC Xilinx Zynq 7100 for the MMP module from AvNet manufacturer. This PSoC is composed of a dual-core ARM-based processing system, and a Kintex-7 FPGA. Both systems are connected using the AXI4-Stream with Direct Memory Access (\textbf{DMA}) that is an open source protocol. It is used to connect the FPGA with its ARM processor, which runs Petalinux operating system (OS) to control the accelerator. It manages the read and write operations between the DDR memory of the AvNet MMP board (outside the PSoC) and the FPGA block-RAM (BRAM) of the accelerator. The processor also computes the last layer of the CNN, called fully-connected layer (FL) after the convolutional layers. It has included in the OS an embedded USB host controller module to interface it to an iniLabs DAVIS240C neuromorphic event-based camera~\cite{Brandli_etal14} for the real-
time demonstrations used to measure the results of this work. For this use, the ARM processor also runs iniLabs cAER\footnote{https://github.com/inivation/caer}, an open source
framework to interface to the DAVIS camera. Fig~\ref{fig:nullhop-axi} shows
the block diagram of the whole FPGA architecture, including the MM2S ($Memory$ $Mapped$ $To$ $Stream$) and S2MM ($Stream$ $to$ $Memory$ $Mapped$) modules used to interface the accelerator with the AXI Stream bus.
MM2S and S2MM include FIFOs for data transfer and finite-state-machines for protocol adaptation between the Xilinx AXI-DMA IP and the NullHop interface, including burst transfer length parameters and burst interruption control.

The design minimizes host memory manipulation by using DMA transfers from host
memory (external DDR connected to the PSoC) to the accelerator and vice versa without requiring each layer output to be reformatted or processed. For each layer, the ARM loads the layer configuration and
the kernels. It then initiates interrupt-driven DMA transfers of the input and output. It is then free for other processing while the layer is computed. 

\begin{figure}[ht!]
\centering
\includegraphics[width=1.0\linewidth]{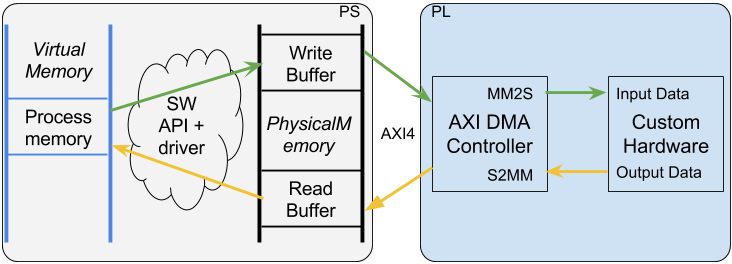}
\caption{DDR Memory hierarchy in a PSoC with OS. User app works at virtual space, while DMA controller at PL works with physical one. The API and/or driver do the transfers to/from both spaces.
\label{virtual_physical_mem}}
\end{figure}

Fig. \ref{virtual_physical_mem} shows the memory hierarchy from the user application to the CNN accelerator. There are two ways to communicate with devices when working with embedded Linux OS: (1) \textit{user-level}: function  \textit{mmap()} for mapping a segment of the device physical address space into our process virtual address space. This function is called by user application directly and the DMA transfers can be configured in a polling scheme, where the user application is frequently blocked, waiting for the transfer to be completed to process the data; or (2) \textit{kernel-level}: a routine running at a higher privilege level of the OS, with interrupt support, maintain the user application free of blocking states until data is ready. Furthermore, the kernel-level ensures the integrity of the software avoiding the possible wrong use of physical address spaces reserved to other processes running in the OS.

This cAER software integration for DVS histogram collection and normalization calculation is requiring a considerable amount of computational resources from ARM processor and power consumption. In this sense, we propose in this work to remove the cAER from the system and the USB-host interface to the DAVIS240c and implement on the FPGA a new block able to take events from a parallel AER DVS retina, collect the histograms and perform the normalization in the same way as it was done in the cAER. 

Next section presents the circuit and results section compares the performances of a real demonstration with both implementations (cAER and FPGA normalizations).

\section{Events Histogram Normalization}

In order to improve performance when using DVS for CNN accelerators beyond 30fps, it has been designed a circuit for FPGA able to collect events from the DVS sensor in a histogram (in BRAM), normalize this histogram using DSP slices, converts it to the sparsity maps format and feed this codded histogram to the \textit{NullHop} accelerator in a real-time continuous way. The normalization to be implemented must be the same used in both the training dataset and in the software version of the example. This normalization responds to the algorithm expressed mathematically at eq. \eqref{eq:norm}.

\begin{equation} \label{eq:norm}
\begin{split}
\begin{array}{c}
    S=\sum_{a=0}^N\sum_{b=0}^N F_{in}(a,b) \\ \\
    c=\sum_{a=0}^N\sum_{b=0}^N f(F_{in}(a,b)) \\ \\ f(X)= \bigg\{
  \begin{tabular}{l}
0 ,if X=0\\ 
1 ,otherwise 
  \end{tabular}
\\ \\
    mean = S/c \\ \\
    \sigma = \sqrt{\frac{\sum_{a=0}^N\sum_{b=0}^N [F_{in}(a,b)-mean]^2}{c}} \\ \\
    F_{norm}(i,j)=\frac{F_{in}(i,j)+3\sigma}{6\sigma}, \forall{i,j\in[0,N]}
\end{array}
\end{split}
\end{equation}

Where $S$ is the accumulation of all non-zero pixel's values, $c$ is the number of pixels with non-zero values, $mean$ is the averaged pixel value from those greater than zero, $\sigma$ is the square-root of the variance, which is used to normalise the pixels of the histogram to be processed by the CNN.
       
\begin{figure}[ht!]
  \centering
  \includegraphics[width=1.0\linewidth, height=4cm]{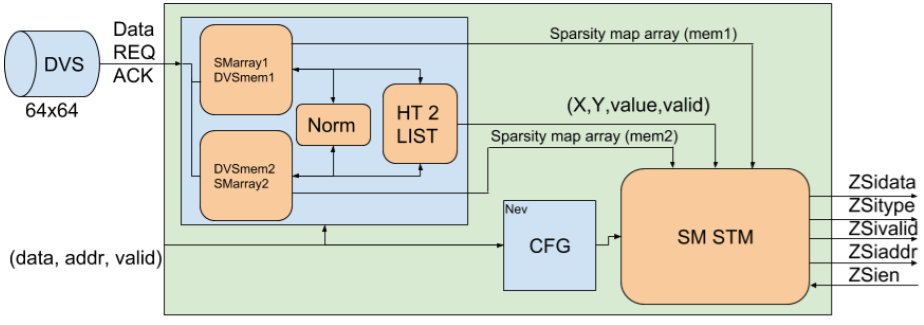}
  \includegraphics[width=1.0\linewidth, height=4cm]{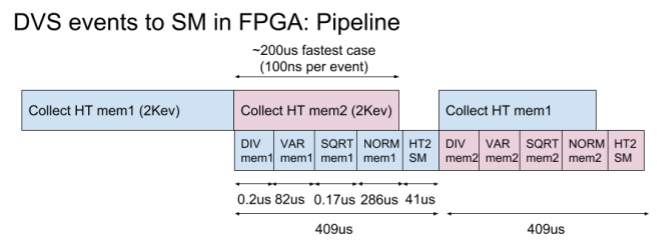}
  \caption{DVS parallel AER events to normalized frame circuit block diagram (top) and pipeline execution scheme (bottom).}
\label{fig:dvs2frm-norm}
\end{figure}

Fig. \ref{fig:dvs2frm-norm} shows the block diagram. Parallel AER DVS events arrive to a double buffered memory (implemented on BRAM) called (DVSmem1, SMarray1) and (DVSmem2, SMarray2). DVSmem stores the histogram of collected events, whilst SMarray stores a mask of those non-zero pixels. A finite state machine (FSM) is configured to collect a number of events (ie. 2K events) in corresponding BRAM as an histogram (HT). When finished, it start to work in the second memory. While the second histogram is being collected, the system keep working on the first histogram, normalizing it according to eq. \eqref{eq:norm} for the CNN example. This normalization requires, in principle, calculations in floating-point domain, and then convert the histogram in a list of non-zero pixels as an intermediate step to generate the sparsity map that NullHop requires as an input.

According to original normalization procedure, several computation units to perform needed operations have been implemented using HLS from Vivado. More precisely, division, variance, square root and the normalization operation itself. Instead of using floating point operations, HLS computation units have been described using the fixed point notations Q24.16\footnote{Qn.m means $n$ bits for the integer part of the number and $m$ bits for the decimal part} for internal operations and Q16.8 for final normalisation output. Once normalization is done in the corresponding BRAM memory, a last state machine converts its content into a sequence of properly codded sparsity maps readable by NullHop CNN accelerator.
It can be seen in Fig. \ref{fig:dvs2frm-norm} the needed time for each HLS module and the last state machine, which is longer than the required time for collecting 2k events when the DVS is sending events at its maximum throughput.

\begin{figure}[ht!]
  \centering
  \includegraphics[width=1.0\linewidth, height=10cm]{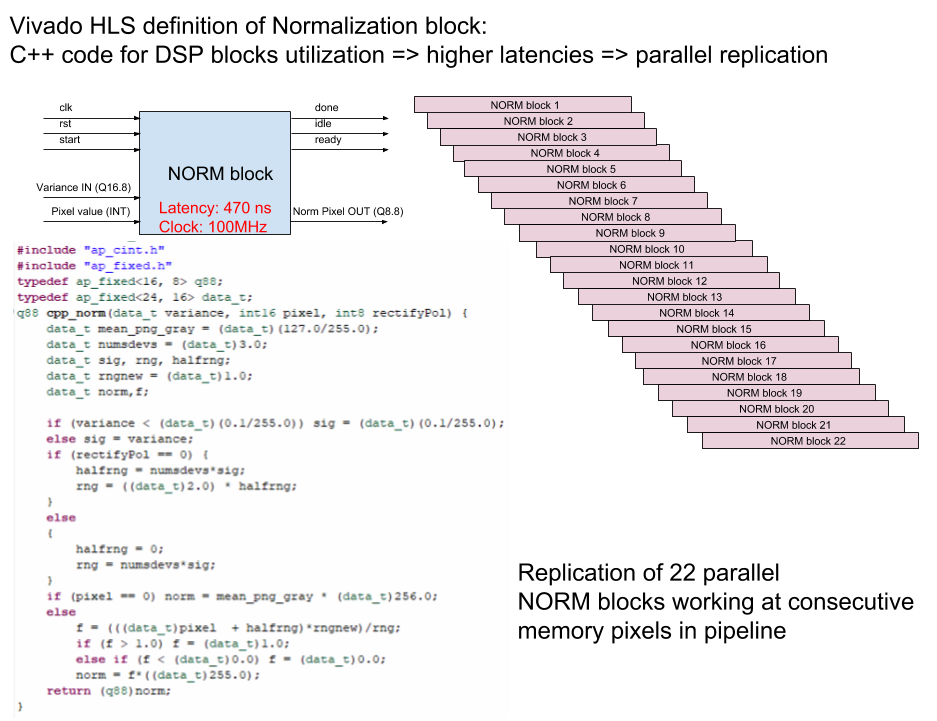}
  \caption{HLS block for NORM stage. It is instantiated 22 times in pipeline.}
\label{fig:norm-hls}
\end{figure}

In order to extract the maximum performance we have implemented as many instances of the HLS blocks as needed to reduce the final latency for a histogram normalization. For example, as shown in Fig. \ref{fig:norm-hls}, the NORM block has been replicated 22 times. These 22 NORM blocks are feed with data coming from BRAM in consecutive clock cycles for consecutive pixels in BRAM in order to obtain also a continuous stream of normalized pixels coming back to BRAM. After the penalty for the first pixel, that corresponds to the latency of a NORM block\footnote{470ns for a 100MHz clock and 783ns for a 60MHz clock}, one result is obtained per clock cycle. This figure also shows the algorithm of the HLS block.

For 64x64 histogram resolutions, down-sampled from the DVS resolution to be used in the Roshambo experiment, half microsecond is needed for normalizing and converting to sparsity map compression and feeding the HullHop. This is around x4 faster than the cAER solution running at ARM cores, as we demonstrate in the results section.

\section{Experiments and results}

For measuring the performance of the proposed solution we have tested the design in an scenario where the histogram normalization was previously done in C++ software (from the caer library) running under Petalinux OS on the ARM cores of a PSoC Xilinx device for classifying hand symbols presented to a DVS retina to play Roshambo game. The used CNN was developed to extract the maximum performance over the NullHop CNN accelerator in terms of MAC utilisation, as described in \cite{Aimar_etal2019}.

\begin{figure}[ht!]
  \centering
  \includegraphics[width=1.0\linewidth]{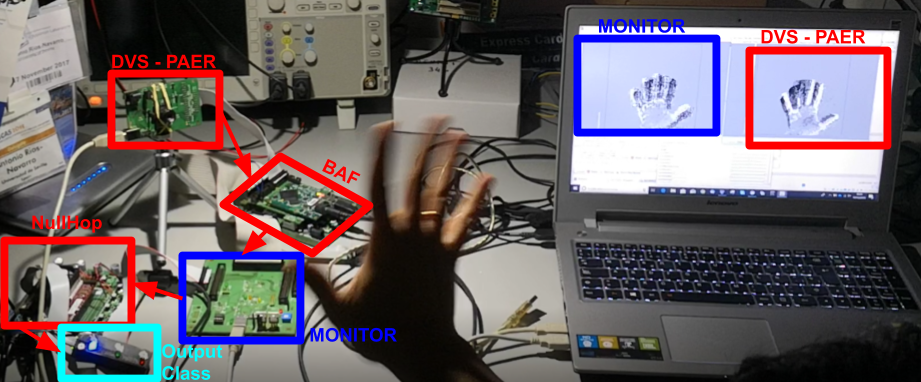}
  \caption{Roshambo game scenario with DVS and NullHop.}
\label{fig:roshambo-scenario}
\end{figure}

Fig. \ref{fig:roshambo-scenario} shows the testing scenario. A DVS retina with parallel AER output is connected to an AER-Node board \cite{Yousefzadeh_etal2017}, plus an OKAERtool \cite{rios_etal2016} running an event-based Background Activity Filter (BAF) \cite{linares_etal2015} and monitoring through USB the output of the DVS, whose histogram can be seen on the right part of the laptop screen. The output of the BAF is connected to a USBAERmini2 board \cite{Berner_etal07}, which monitor the stream of events through USB in the left part of the laptop screen, and at the same time, keep sending monitored events to the MMP module that is running Roshambo on the NullHop CNN accelerator. The output of the NullHop is sent to the LED's bar to highlight the winner class. In this picture, the hand in motion presented to the DVS is in "paper" gesture, and the LED that is turned ON is the "scissors" gesture, thus the machine wins the game.

The CNN for playing Roshambo consists of 5 convolution layers and a fully-connected-layer executed in the ARM cores. Each convolution layer can be executed at once in the CNN accelerator after downloading the corresponding layer parameters and the input feature maps. Therefore, the software controller running on the ARM cores only needs to iterate 5 times with the accelerator to write and read each feature maps and the layer parameters using the designed AXIDMA interface to extract the maximum performance of $\approx 8ms$ per frame as described in \cite{rios_etal18_nano}.

This software controller of the ARM cores has been modified to allow the system to receive the input feature maps for the first layer from the circuit presented in the previous section, instead of receiving it from the AXIDMA. Therefore, with this solution, there is no need of cAER library utilisation, neither USB-host interface running at ARM cores. This decreases the power consumption.

\begin{figure}[ht!]
  \centering
  \includegraphics[width=1.0\linewidth, height=15cm]{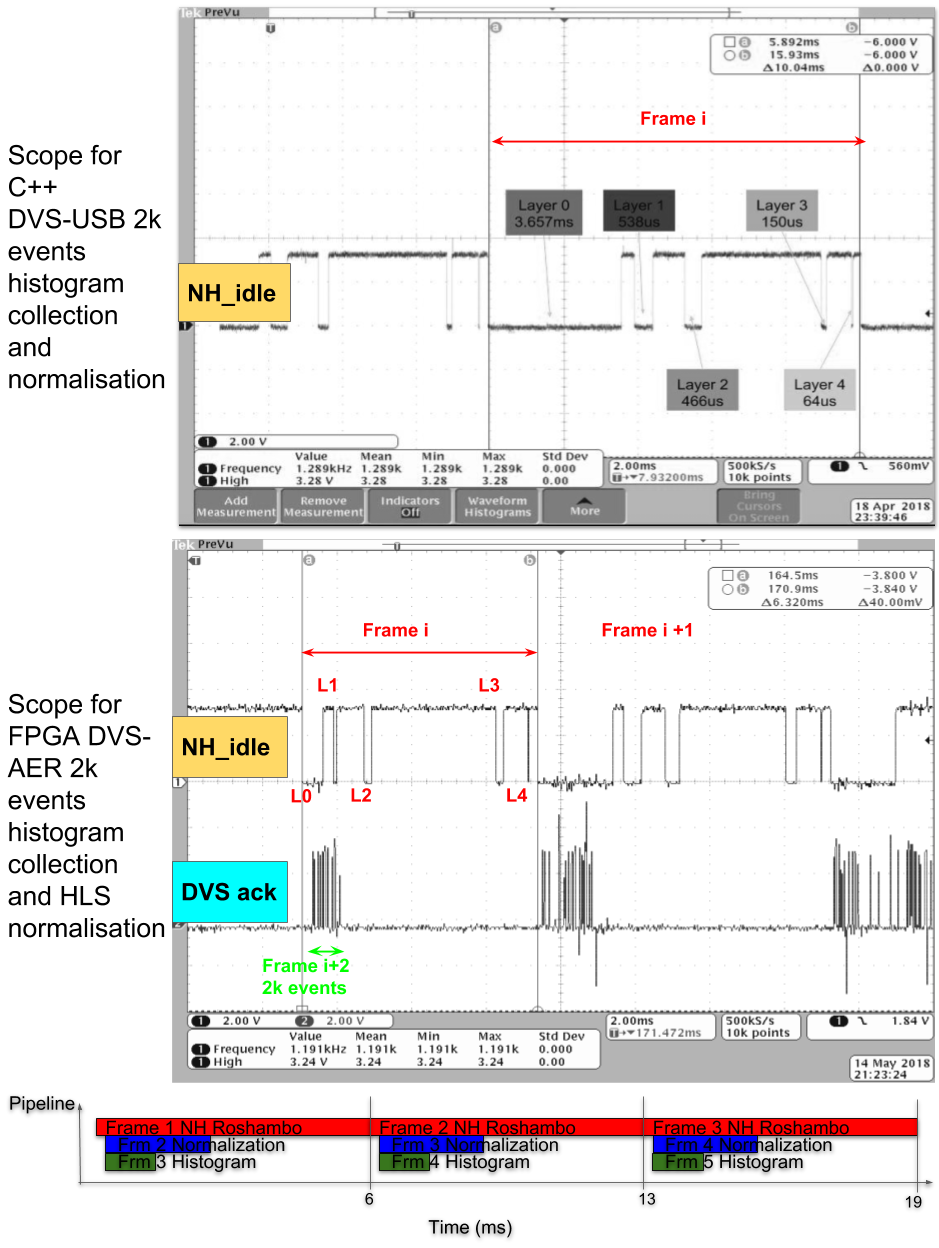}
  \caption{Scope captures for cAER histogram capture / normalisation (top) vs FPGA HLS normalisation (middle) running Roshambo CNN on NullHop. Pipelining performance improvement (bottom).}
\label{fig:caer-hls}
\end{figure}

Fig. \ref{fig:caer-hls} shows two oscilloscope captures of NullHop solving the Roshambo CNN at 60MHz connected to ARM cores running at 800MHz. Top capture shows the behavior when using cAER and DVS retina connected through USB, where normalisation is done in software. \textit{NH\_idle} signals if the accelerator is working ('0') or idle ('1'). It is highlighted the sequence along the five convolutional layers. Running at real-time and playing Roshambo game, the normal latency from frame to frame measured was $\approx 10ms$. Bottom capture shows the behavior when the events histogram collection and normalisation is done on the FPGA for the same number of events (2k) per histogram. It can be seen how the DVS activity is stopped if the sensor is stimulated with a high speed object. In this case, the minimum measured time per frame's classification has been $\approx6ms$. This implies a 67\% speed up factor. 
Furthermore, when using this faster solution, the system has been improved by allowing pipelined behaviour between the stages of (1) events collection, (2) histogram normalization and (3) Roshambo computation on the NullHop accelerator; as shown in the bottom of the Fig. \ref{fig:caer-hls}, where it is highlighted how normalised frames wait on BRAM ready to be sent for computation while the previous frame is being computed by Roshambo CNN. 

\begin{table}[]
\caption {FPGA resource consumption for NullHop, DVS normalisation (DVS2SM), AXIDMA and debugging logic.}
\centering
\begin{tabular}{|l|c|c|c|c|}
\hline
\rowcolor[HTML]{3531FF} 
{\color[HTML]{FFFFFF} \begin{tabular}[c]{@{}c@{}}Zynq7100\\ (\#)\end{tabular}} & \multicolumn{1}{c|}{\cellcolor[HTML]{3531FF}{\color[HTML]{FFFFFF} 
\begin{tabular}[c]{@{}c@{}}LUTs \\ (277K)\end{tabular}}} & 
\multicolumn{1}{c|}{\cellcolor[HTML]{3531FF}{\color[HTML]{FFFFFF} 
\begin{tabular}[c]{@{}c@{}}FlipFlops \\ (555K)\end{tabular}}} & 
\multicolumn{1}{c|}{\cellcolor[HTML]{3531FF}{\color[HTML]{FFFFFF} 
\begin{tabular}[c]{@{}c@{}}BlockRAM \\ (755)\end{tabular}}} & \multicolumn{1}{c|}{\cellcolor[HTML]{3531FF}{\color[HTML]{FFFFFF}
\begin{tabular}[c]{@{}c@{}}DSP \\ (2020)\end{tabular}}} \\ \hline
\rowcolor[HTML]{DAE8FC} 
\cellcolor[HTML]{3166FF}{\color[HTML]{FFFFFF} RTC\_DVS2SM} & \textbf{50} & \textbf{25} & \textbf{4} & \textbf{513} \\ \hline
\rowcolor[HTML]{DAE8FC} 
\cellcolor[HTML]{3166FF}{\color[HTML]{FFFFFF} NullHop} & \textbf{210} & \textbf{107} & \textbf{385} & \textbf{144} \\ \hline
\rowcolor[HTML]{DAE8FC} 
\cellcolor[HTML]{3166FF}{\color[HTML]{FFFFFF} AXIDMA} & \textbf{5} & \textbf{6} & \textbf{11} & \textbf{0} \\ \hline
\rowcolor[HTML]{DAE8FC} 
\cellcolor[HTML]{3166FF}{\color[HTML]{FFFFFF} Debug} & \textbf{1} & \textbf{1} & \textbf{0} & \textbf{0} \\ \hline
\rowcolor[HTML]{34CDF9} 
\cellcolor[HTML]{3531FF}{\color[HTML]{FFFFFF} Total} & \textbf{266} & \textbf{139} & \textbf{401} & \textbf{657} \\ \hline
\end{tabular}
\label{table:fpga_resources}
\end{table}

In Table \ref{table:fpga_resources} can be seen in the first row the available resources for a Zynq 7100 PSoC (from Xilinx) regarding to LUTs, flip-flops slices, BRAM and DSP slices. Following rows present the consumed resources for each part of the system. It can be seen how the use of HLS for implementing the normalisation has considerably increased the number of DSP slices compared to the previous version, where only 144 DSP slices were needed to implement the 128 MACs and other side operations in the accelerator. There is a reasonable increment on the number of resources for LUT, FF and BRAM for this normalisation circuit.

The power consumption for Zynq chip running the cAER version was of 272mW for the logic and it was increased up to 1W when the ARM cores were executing the Roshmabo game. For the hardware version presented in this paper, the total power consumption running the same game was of 500mW, including the ARM cores.

\section{CONCLUSIONS}

Neuromorphic vision sensors, such as DVS, on deep-learning applications allow to surpass the limit imposed by conventional commercial cameras of 30fps. There are numerous applications where faster than 30fps video sources are required to be processed. Autonomous driving would need specialised hardware able to detect pedestrian, cars, traffic signs, obstacles, ..., faster than these 30fps to guarantee the efficiency of the system. Powerful deep-learning CNN algorithms need to compute magnificent amount of operations, what makes it still inviable to go beyond 30fps even with GPU based solutions. More specialized and simpler CNNs are appearing to mitigate this problem. Therefore, by replicating as many times as needed the number of CNNs it will be possible to obtain faster results and in a parallel way.
We propose to use DVS retinas as a vision source for CNN accelerators to be able to increase the fps to the limit of the accelerator itself. In this paper, we have demonstrated that using powerful FPGA design technology (ie. HLS) running at 60MHz, a deep-learning system can be speed up a 67\% respect to its software competitor running at 800MHz on embedded ARM cores, with the penalty of an extra use of 18\% LUTs, 4.5\% FF, 0.5\% BRAM and 25\% DSP of the PSoC Zynq 7100 chip.

\addtolength{\textheight}{-12cm}   % This command serves to balance the column lengths

%%%%%%%%%%%%%%%%%%%%%%%%%%%%%%%%%%%%%%%%%%%%%%%%%%%%%%%%%%%%%%%%%%%%%%%%%%%%%%%%

%%%%%%%%%%%%%%%%%%%%%%%%%%%%%%%%%%%%%%%%%%%%%%%%%%%%%%%%%%%%%%%%%%%%%%%%%%%%%%%%

\section*{ACKNOWLEDGMENT}

This work is supported by the excellence project from the Spanish government grant (with support from the European Regional Development Fund) COFNET (TEC2016-77785-P); and by SEC project NPP  (P051-15/E03) coordinated by Prof. Delbruck. Mr. Ricardo Tapiador-Morales work has  been  supported  by  a "Formación de Personal Investigador" Scholarship from the University of Seville.

\bibliographystyle{IEEEtran}
\bibliography{name.bib}

\end{document}